\documentclass[letterpaper, 10 pt, conference]{ieeeconf}  

\IEEEoverridecommandlockouts                              

\usepackage{graphics} 
\usepackage{epsfig} 
\usepackage{mathptmx} 
\usepackage{times} 
\usepackage{amsmath} 
\usepackage{amssymb}  

\usepackage{booktabs}
\usepackage{multirow}
\usepackage{cite}
\usepackage[table]{xcolor}
\usepackage{hyperref} 
\usepackage{siunitx}
\usepackage{mathtools}

\definecolor{best}{HTML}{FBD5D5}    
\definecolor{second}{HTML}{FBE5C6}  

\title{\LARGE \bf
SVR-GS: Spatially Variant Regularization for Probabilistic Masks in 3D Gaussian Splatting}

\author{%
Ashkan Taghipour$^{1}$,
Vahid Naghshin$^{2}$,
Benjamin Southwell$^{2}$,
Farid Boussaid$^{3}$,
Hamid Laga$^{4}$,
Mohammed Bennamoun$^{1}$%
\thanks{This work was done while Ashkan Taghipour was a research intern at Dolby Laboratories, Sydney, Australia.}%
\\[2pt]
$^{1}$Department of Computer Science and Software Engineering, The University of Western Australia, Australia.\\
$^{2}$Dolby Laboratories, Sydney, Australia.\\
$^{3}$Department of Electrical, Electronics and Computer Engineering, The University of Western Australia, Australia.\\
$^{4}$School of Information Technology, Murdoch University, Australia.\\[4pt]
\texttt{ashkan.taghipour@research.uwa.edu.au}\\
\texttt{vahid.naghshin@dolby.com, benjamin.southwell@dolby.com}\\
\texttt{farid.boussaid@uwa.edu.au, h.laga@murdoch.edu.au, mohammed.bennamoun@uwa.edu.au}
}

\begin{document}

\maketitle
\thispagestyle{empty}
\pagestyle{empty}

\begin{abstract}
3D Gaussian Splatting (3DGS) enables fast, high-quality novel view synthesis but relies on densification followed by pruning to optimize the number of Gaussians. Existing mask-based pruning, such as MaskGS, regularizes the \emph{global} mean of the mask, which is misaligned with the \emph{local} per-pixel (per-ray) reconstruction loss that determines image quality along individual camera rays. This paper introduces \textbf{SVR-GS}, a spatially variant regularizer that renders a per-pixel spatial mask from each Gaussian’s effective contribution along the ray, thereby applying sparsity pressure where it matters: on low-importance Gaussians. We explore three spatial-mask aggregation strategies, implement them in CUDA, and conduct a gradient analysis to motivate our final design. Extensive experiments on \textit{Tanks\&Temples}, \textit{Deep Blending}, and \textit{Mip-NeRF360} datasets demonstrate that, on average across the three datasets, the proposed \textbf{SVR-GS} reduces the number of Gaussians by \textbf{1.79}\,\scalebox{1.2}{$\times$} compared to MaskGS and \textbf{5.63}\,\scalebox{1.2}{$\times$} compared to 3DGS, while incurring only \textbf{0.50}\,\text{dB} and \textbf{0.40}\,\text{dB} PSNR drops, respectively. These gains translate into significantly smaller, faster, and more memory-efficient models, making them well-suited for real-time applications such as robotics, AR/VR, and mobile perception. The code will be released upon publication.
\end{abstract}

\section{Introduction}

Novel view synthesis (NVS) aims to render realistic images of a 3D scene from camera poses that were never observed \cite{triangle_splat, he2025survey, convex, tum, superspalt}. Robotic autonomy relies on accurate 3D reconstruction, and novel view synthesis (NVS) complements it as a predictive sensor by rendering expected views from candidate poses. These renderings expose visibility and occlusions, improving next-best-view selection and safer planning \cite{zhou2023nerf}. In practice, this supports navigation~\cite{navigation}, mapping~\cite{maping}, manipulation~\cite{manip}, and active perception~\cite{perception}.
A major step forward in NVS was the introduction of Neural Radiance Fields (NeRF)~\cite{nerf,Lin_2025_CVPR}, which represent a scene as a continuous function queried by a multi-layer perceptron (MLP).
While NeRFs yield photorealistic rendering, they are computationally intensive~\cite{qi2023e2nerf,qu2024lush}, typically requiring \(\sim 150\text{--}200\) MLP evaluations per pixel (e.g., 64 coarse + 128 fine), making them unsuitable for real-time deployment in robotics~\cite{icra_edgs}---particularly under the constraints of limited compute and memory on embedded platforms.
3D Gaussian Splatting (3DGS) replaces numerous per-sample MLP queries with an explicit set of anisotropic 3D Gaussians~\cite{3dgs_orig}. At render time, each 3D Gaussian is projected to a 2D elliptical Gaussian footprint (a “splat”) and composited front-to-back with alpha blending~\cite{survey_2}. The pipeline runs efficiently on GPUs by grouping splats into \(16\times16\)-pixel tiles and compositing them in shared memory, avoiding per-ray MLP evaluations~\cite{ICRA_2,spotless}.

However, constructing 3DGS representations typically combines densification, parameter optimization, and pruning~\cite{miniSplating,zhang2025cobgs,3dgs_optim}.
Densification adds new Gaussians to improve coverage and detail by splitting large Gaussians, cloning high-gradient Gaussians, and seeding under-covered regions~\cite{densification,densification_2}, often yielding scenes with millions of Gaussians.
Pruning subsequently removes Gaussians with minimal impact on the photometric reconstruction of the training views, thereby reducing memory usage and accelerating training and inference-time rendering~\cite{compressed,contextgs,compact}.

In this context, recent work has focused on reducing the Gaussian count while preserving image quality \cite{radsplat, footprint, lp3dgs}.
Building on this trend, Compact3DGS~\cite{compact3dgs} learns a binary mask per-Gaussian that gates both opacity and scale. Using a straight-through estimator, it hard-masks tiny or transparent Gaussians whose contribution is visually negligible, reducing the number of low-contribution Gaussians during training. Rate--Distortion Optimization (RDO)~\cite{rdo_gaussian} learns a per-Gaussian mask on opacity and scale. During training, the masks remain soft and dynamic: Gaussians may be temporarily suppressed and later revived. Hard removal is applied only after optimization.

LightGaussian~\cite{lightgaussian} assigns an importance score to each Gaussian based on its contribution across training views, and prunes those with consistently low scores.
EfficientGS~\cite{efficientgs} streamlines 3DGS by improving densification with aggregated pixel-gradient magnitudes, densifying only Gaussians that remain non-steady and thereby avoiding unnecessary splits. For pruning, it adopts an opacity-based criterion to remove faint Gaussians. Reduced 3DGS~\cite{reduce_3dgs} detects regions where Gaussians are densely packed and prunes those that heavily overlap with their neighbors to avoid overpopulation.
PUP--3DGS~\cite{pup} estimates a sensitivity score from the reconstruction-error Hessian and prunes Gaussians that show high spatial uncertainty and low impact on image quality. Taming 3DGS~\cite{taming} and RN-Gaussian~\cite{rn_gaussian} limit the cloning and splitting of Gaussians by considering their position and correlation with nearby Gaussians, thereby controlling growth and reducing redundancy.

\begin{figure}[!t]
  \centering
  \includegraphics[width=\columnwidth]{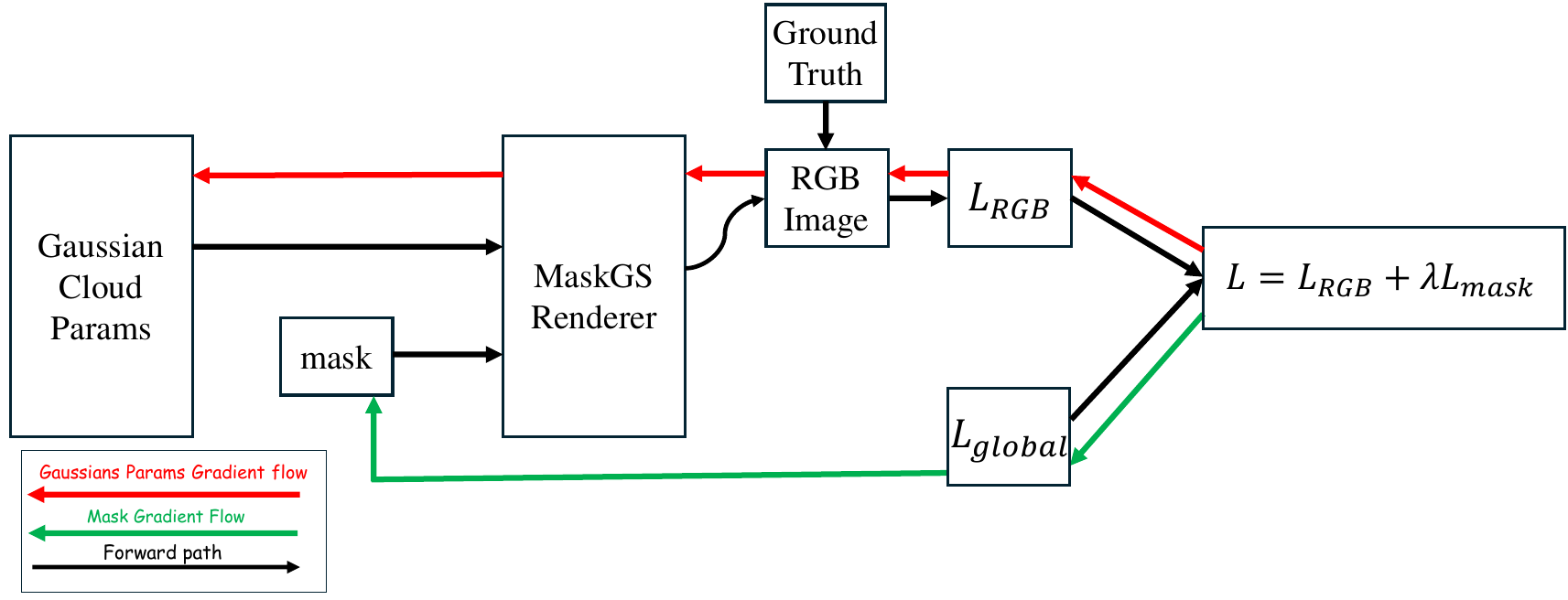} 
\caption{MaskGS learns per-Gaussian existence probabilities and samples a binary mask to gate contributions while keeping Gaussian parameters trainable. However, its global mean-mask regularizer (Eq.\ref{eq:global-mask-loss}) enforces uniform sparsity, ignoring per-ray importance and motivating our spatially variant objective.}

\label{fig:maskgs_flow}
\end{figure}

MaskGS~\cite{maskgs} models pruning as learning a probability of existence for each Gaussian. 
During training, it constructs a binary mask by sampling this probability.
This mask dictates whether a Gaussian contributes to the rendered image without destroying (zeroing) the Gaussian’s parameters (e.g., position, covariance, opacity, color), which remain unchanged and trainable. 
The Gaussian's mask probabilities are regularized by adding the square of the mean Gaussian mask probability to the loss.
Gradients flowing from the reconstruction loss to the Gaussian's that can improve the reconstruction quality result in the mask probabilities increasing (see Fig.~\ref{fig:maskgs_flow}).
Pruning is applied at each densification step up to iteration \(15\mathrm{k}\), and thereafter every \(1{,}000\) iterations until \(30\mathrm{k}\); Gaussians are removed based on repeated stochastic sampling of their existence probabilities.
As a result, the regularization term results in fewer Gaussians in the final optimized Gaussian cloud.
Whilst achieving SOTA reconstruction quality on standard metrics and using fewer Gaussians than prior methods, MaskGS applies a uniform mask regularization that does not account for each Gaussian’s contribution to the rendered images.

In this paper, we propose SVR-GS, a spatially variant regularization method for the probabilistic mask scheme proposed in MaskGS~\cite{maskgs}.
We do this by constructing a regularization function that considers the importance of each Gaussian in the RGB image by rendering a spatial mask image to regularize rather than simply taking the square of the mean probabilities.
The spatial mask image penalizes each Gaussian according to its contribution, or more specifically, lack thereof, to pixels in the RGB image. 
This improves upon the MaskGS approach as we are now penalizing the existence of a Gaussian according to its contribution to the rendered images directly.
The contributions of this paper can be summarized as:
\begin{itemize}

    \item \textbf{Spatially variant mask objective.} A local, per-pixel regularizer on each Gaussian’s \emph{probability of existence} (mask probability), with weights derived from its visibility-weighted contribution to that pixel. This selectively penalizes Gaussians that contribute little to the rendered RGB image while preserving high-impact ones, aligning the pruning signal with the photometric reconstruction objective.

    \item \textbf{Design space analysis of spatial-mask renderers.} Three forward spatial mask rendering designs are implemented in CUDA and analyzed to explain their behavior dynamics; the ablation study reveals how design choices shape mask quality and reconstruction, motivating the final formulation.

\item \textbf{Empirical effectiveness.} Comprehensive experiments on three real-world datasets show that, on average across the three datasets, our regularizer reduces the number of Gaussians by \textbf{1.79}\,$\boldsymbol{\times}$ compared to MaskGS and \textbf{5.63}\,$\boldsymbol{\times}$ compared to 3DGS, with PSNR drops of less than \textbf{0.5}~dB (\textbf{0.50} and \textbf{0.40}~dB), yielding substantially smaller models with negligible impact on reconstruction quality.

\end{itemize}

\begin{figure*}[t] 
    \centering
    \includegraphics[width=\textwidth]{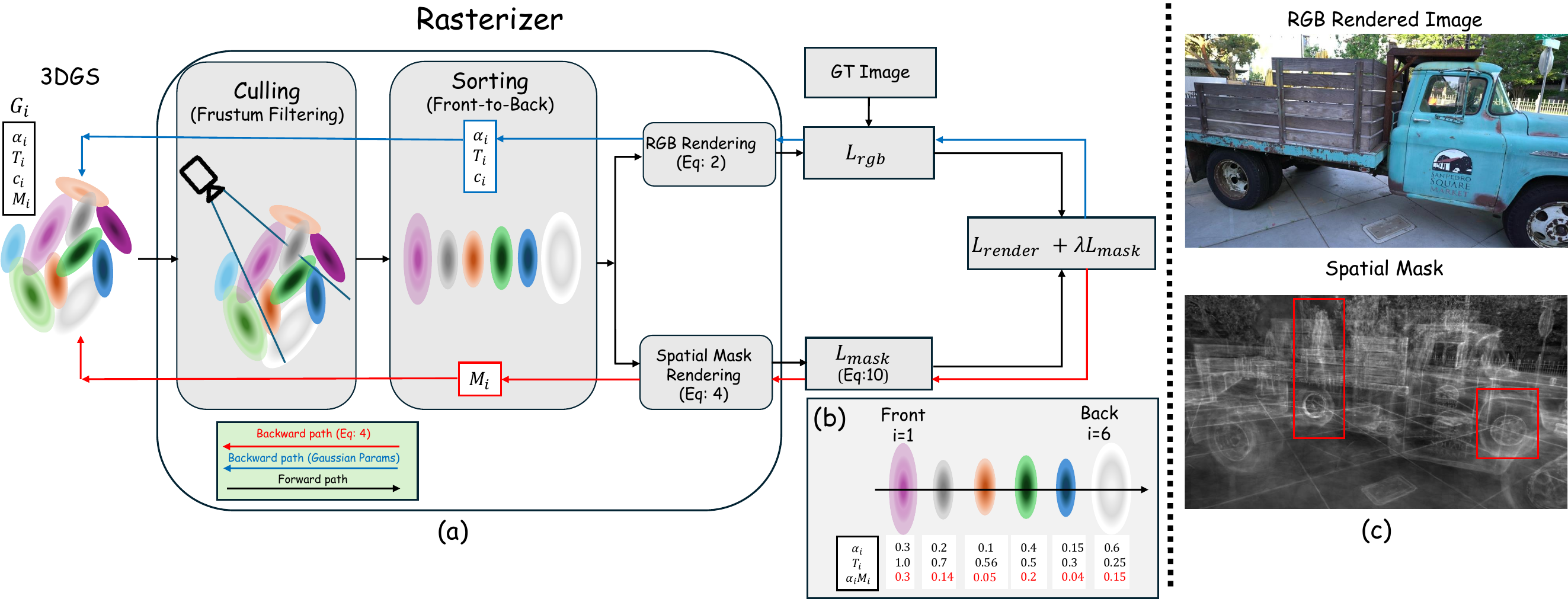}
\caption{
\textbf{Pipeline overview.}
(a) Starting from a set of 3D Gaussians $G_i$, Rasterizer cull splats outside the camera frustum and sort the remainder front-to-back. The renderer then branches into: (i) an RGB Rendering path that composites colors using Eq.~\eqref{eq:color-compositing}; and (ii) a Spatial Mask rendering path that renders a Spatial Mask from the masks $M_i$ (Eq.~\eqref{eq:redundancy-map}). The training objective combines the image loss $L_{\text{rgb}}$ with the mask loss as $L_{\text{render}}+\lambda L_{\text{mask}}$ (Eq.~\eqref{eq:total-loss}). Blue and red arrows denote gradient flow from the RGB and mask losses, respectively; black arrows indicate the forward pass. (b) Front-to-back ordering example ($i{=}1$ is nearest); the table lists per-splat opacity $\alpha_i$, transmittance $T_i$, and masked contribution $\alpha_i M_i$. (c) Example outputs: RGB rendering (top) and the Spatial Mask (bottom), with regions highlighted (red boxes) where redundancy is emphasized.}
    \label{fig:main_double}
\end{figure*}

\section{Proposed Method}

\subsection{Preliminaries: MaskGS}
\label{subsec:preliminaries_maskgs}
MaskGS~\cite{maskgs} optimizes a mask probability for each Gaussian and uses a differentiable Gumbel--Softmax~\cite{gumble} during training. Masks do not modify a Gaussian’s intrinsic attributes (e.g., opacity, shape, color); they simply decide whether it participates in compositing. Gaussians intersecting the ray for pixel \(x\) are composited in front-to-back order, with \(i{=}0\) the nearest. Each Gaussian has spherical harmonics (SH) appearance coefficients \(\mathbf{c}_i\); given the viewing direction \(\mathbf{v}(x)\) at pixel \(x\), its view-dependent color is \(c_i(\mathbf{v}(x))\). The projected 2D elliptical footprint determines the per-pixel opacity \(\alpha_i(x)\). The accumulated transmittance \(T_i(x)\) is the fraction of light remaining before blending Gaussian \(i\) (with \(T_0(x)=1\)). The
transmittance  and rendered color are then given by

\begin{equation}
T_{i+1}(x) \;=\; \bigl(1 - M_i\,\alpha_i(x)\bigr)\,T_i(x),
\label{eq:transmittance-update}
\end{equation}

\begin{equation}
c(x) \;=\; \sum_{i=0}^{N-1} M_i\,\alpha_i(x)\,c_i\!\bigl(\mathbf{v}(x)\bigr)\,T_i(x).
\label{eq:color-compositing}
\end{equation}

\noindent Thus, \(M_i{=}1\) includes Gaussian \(i\) (adds its color term and reduces transmittance); \(M_i{=}0\) excludes it from Eq.~\ref{eq:color-compositing} and leaves \(T_{i+1}(x)=T_i(x)\) unchanged. So, the mask only decides whether a Gaussian participates in compositing; it does not delete or zero the Gaussian’s parameters, which remain in the model and continue to be optimized. 
The overall objective combines a standard per-pixel reconstruction loss,
\(L_{\text{rgb}} = L_{1} + (1-\mathrm{SSIM})\), where \(L_{1}\) is the mean absolute error and SSIM is the structural similarity index measure, with a sparsity term on the average mask probability:

\begin{equation}
L \;=\; L_{\text{rgb}} \;+\; 
\underbrace{\lambda_m \left(\frac{1}{N}\sum_{i=0}^{N-1} M_i\right)^{2}}_{L_{\text{global}}},
\label{eq:global-mask-loss}
\end{equation}

\noindent where \(N\) is the number of Gaussians and \(\lambda_m>0\) controls the regularization strength.
Pruning is stochastic: at each pruning event, MaskGS~\cite{maskgs} sample each Gaussian’s mask \(10\) times and remove Gaussians that never activate; this is applied at every densification step and, thereafter, at fixed intervals of \(1{,}000\) iterations.
A key limitation is that A single global average (second term of Eq.~\ref{eq:global-mask-loss}) treats all Gaussians equally, regardless of whether they are critical along a few rays or effectively invisible due to occlusion. So the penalty pushes down both indiscriminately. As a result, the optimizer may suppress high-impact Gaussians that support fine detail while under-penalizing the ones that have low contributions, leading to either quality loss or insufficient pruning.

\subsection{Spatial Mask Rendering}
\begin{figure}[t] 
  \centering
  \includegraphics[width=0.9\columnwidth]{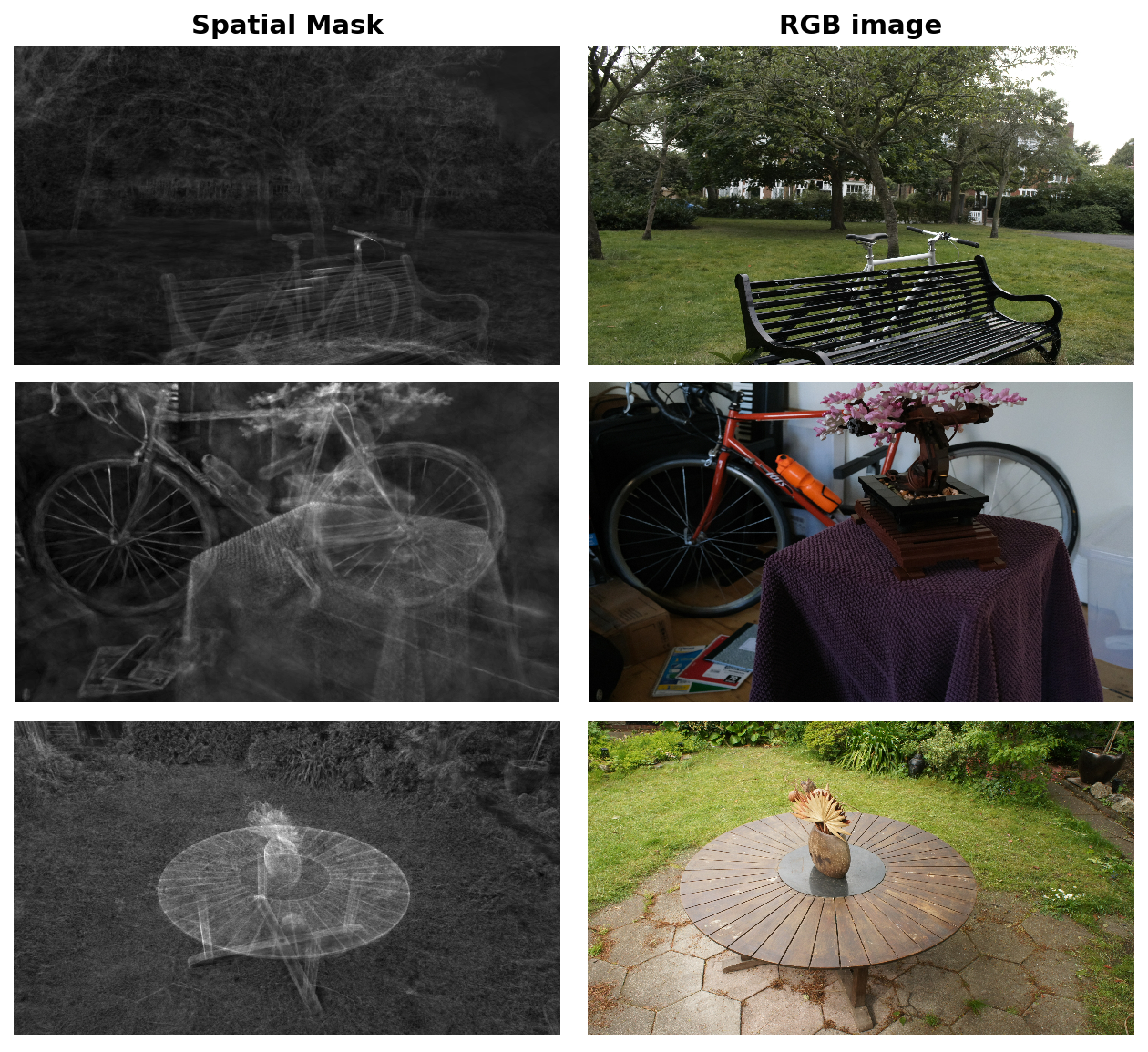}
    \caption{Spatial mask \(F(x)\) rendered via Eq.~\eqref{eq:redundancy-map} (left) alongside the RGB image (right). High intensities correspond to small per-Gaussian contributions \(\alpha_i(x)T_i(x)\), highlighting occluded or faint structures.}

  \label{fig:forward_visual_figure}
\end{figure}

Our goal is a local, per-pixel spatial mask \(F(x)\) that highlights \emph{low-importance} Gaussians along each camera ray, namely those that are low-opacity or occluded. We use the per-Gaussian importance signal \(\alpha_i(x)\,T_i(x)\). This quantity is large for front-most, high-opacity Gaussians with high transmittance \(T_i(x)\), and it is small for occluded or low-opacity Gaussians. We invert this importance to score low-importance and aggregate it along the ray:

\begin{equation}
\label{eq:redundancy-map}
F(x) \;=\; \frac{1}{\log\!\bigl(1+N(x)\bigr)}
\sum_{i=0}^{N(x)-1}
M_i\,\Bigl(1-\alpha_i(x)\,T_i(x)\Bigr),
\end{equation}

\noindent where \(N(x)\) denotes the number of Gaussians intersecting the ray through pixel \(x\), \(M_i\in\{0,1\}\) is the sampled binary mask, and \(T_i(x)\) is the transmittance prior to blending Gaussian \(i\) (updated via Eq.~\ref{eq:transmittance-update}), with initialization \(T_0(x)=1\).
 The factor \(\log\!\bigl(1+N(x)\bigr)\) normalizes the scale across rays of different lengths. Intuitively, \(F(x)\) becomes large when the ray contains many \emph{low-importance} Gaussians (good pruning targets) and remains small when a few foreground Gaussians explain most of the rendered color \(c(x)\) at that pixel (see Fig.~\ref{fig:forward_visual_figure}.

\subsection{Backward: Gradient of the Mask Regularizer}
\label{sec:backward}

For a given pixel \(x\), we order the Gaussians along the ray by depth: \(i=0\) is nearest, then \(i=1,\dots,N{-}1\).
We differentiate \(F\) with respect to a fixed mask \(M_i\). We use \(j\) as a running index in sums and \(k\) as a dummy index in products. For clarity of presentation, we omit the pixel index \(x\).

We write the transmittance as
\begin{equation}
T_j \;=\; \prod_{k<j}\bigl(1-\alpha_k M_k\bigr), \qquad T_0=1.
\label{eq:transmittance}
\end{equation}
With this setup, \(\partial F/\partial M_i\) has two parts: a direct effect on the \(i\)-th term and an indirect occlusion effect on later Gaussians.

Let \(f_i = M_i\bigl(1-\alpha_i T_i\bigr)\), so \(F = \tfrac{1}{\log(1+N)}\sum_i f_i\).

\noindent\textbf{Self term (index \(i\)).}\;
Since \(T_i\) depends only on masks in front (\(k<i\)), it does not depend on \(M_i\):
\begin{equation}
\frac{\partial f_i}{\partial M_i} \;=\; 1-\alpha_i T_i.
\label{eq:self_term}
\end{equation}

\noindent\textbf{Occlusion term (indices \(j>i\)).}\;
For \(j>i\), \(f_j=M_j(1-\alpha_j T_j)\) depends on \(M_i\) only through \(T_j\).
Using \(T_j=\prod_{k<j}(1-\alpha_k M_k)\), only the factor \((1-\alpha_i M_i)\) depends on \(M_i\). Thus,
\begin{equation}
\frac{\partial T_j}{\partial M_i}
= \Biggl(\prod_{\substack{k<j\\k\neq i}} (1-\alpha_k M_k)\Biggr)(-\alpha_i)
= -\,\alpha_i\,\frac{T_j}{1-\alpha_i M_i}.
\label{eq:dTj_dMi}
\end{equation}
By the chain rule,
\begin{equation}
\begin{aligned}
\frac{\partial f_j}{\partial M_i}
&= \frac{\partial f_j}{\partial T_j}\,\frac{\partial T_j}{\partial M_i} \\
&= (-M_j\alpha_j)\!\left(-\alpha_i\,\frac{T_j}{1-\alpha_i M_i}\right)
= M_j\,\alpha_j\,\alpha_i\,\frac{T_j}{1-\alpha_i M_i}.
\end{aligned}
\label{eq:dfj_dMi}
\end{equation}

\noindent Summing both parts and carrying the constant \(1/\log(1+N)\) gives
\begin{equation}
\frac{\partial F}{\partial M_i}
= \frac{1}{\log(1+N)}\left[(1-\alpha_i T_i)
+ \frac{\alpha_i}{1-\alpha_i M_i}\!
  \sum_{j>i} M_j\,\alpha_j\,T_j \right].
\label{eq:dF_dMi}
\end{equation}


\begin{table*}[t]
  \centering
  \setlength{\tabcolsep}{6pt}
  \renewcommand{\arraystretch}{1.15}
\caption{Quantitative comparison on three datasets; best results are highlighted. \#GS denotes millions of Gaussians; ↑ higher is better, ↓ lower is better.}
\label{tab:cross_dataset_comparison}
  \resizebox{\textwidth}{!}{%
  \begin{tabular}{l | c c c c | c c c c | c c c c}
    \toprule
    Dataset &
      \multicolumn{4}{c|}{Tanks\&Temples} &
      \multicolumn{4}{c|}{Deep Blending} &
      \multicolumn{4}{c}{Mip-NeRF360} \\
    \cmidrule(lr){2-5}\cmidrule(lr){6-9}\cmidrule(lr){10-13}
    Metrics &
      PSNR$\uparrow$ & SSIM$\uparrow$ & LPIPS$\downarrow$ & \#GS$\downarrow$ &
      PSNR$\uparrow$ & SSIM$\uparrow$ & LPIPS$\downarrow$ & \#GS$\downarrow$ &
      PSNR$\uparrow$ & SSIM$\uparrow$ & LPIPS$\downarrow$ & \#GS$\downarrow$ \\
    \midrule
    3DGS (SIGGRAPH 23) &
      23.62 & 0.847 & \cellcolor{best}0.176 & 1.825 &
      29.54 & \cellcolor{best}0.905 & \cellcolor{best}0.244 & 2.815 &
      \cellcolor{best}27.51 & 0.811 & \cellcolor{best}0.224 & 3.204 \\
    LightGaussian (NIPs 24) &
      \cellcolor{best}23.76 & 0.843 & 0.187 & 0.626 &
      29.55 & 0.902 & 0.250 & 0.959 &
      27.50 & 0.809 & 0.232 & 1.084 \\
    Compact3DGS (CVPR 24) &
      23.68 & \cellcolor{best}0.849 & 0.179 & 0.960 &
      29.63 & 0.903 & 0.245 & 1.310 &
      27.34 & \cellcolor{best}0.812 & 0.233 & 1.533 \\ 
    MaskGS (CVPR 25) &
      23.75 & 0.847 & 0.178 & 0.590 &
      \cellcolor{best}29.71 & \cellcolor{best}0.905 & 0.246 & 0.694 &
      27.49 & 0.811 & 0.228 & 1.205 \\
    Ours &
      23.25 & 0.829 & 0.221 & \cellcolor{best}0.272 &
      29.46 & 0.904 & 0.261 & \cellcolor{best}0.255 &
      26.75 & 0.800 & 0.255 & \cellcolor{best}0.866 \\
    \bottomrule
  \end{tabular}}
\end{table*}

\begin{figure*}[t] 
  \centering
  \includegraphics[width=0.98\textwidth]{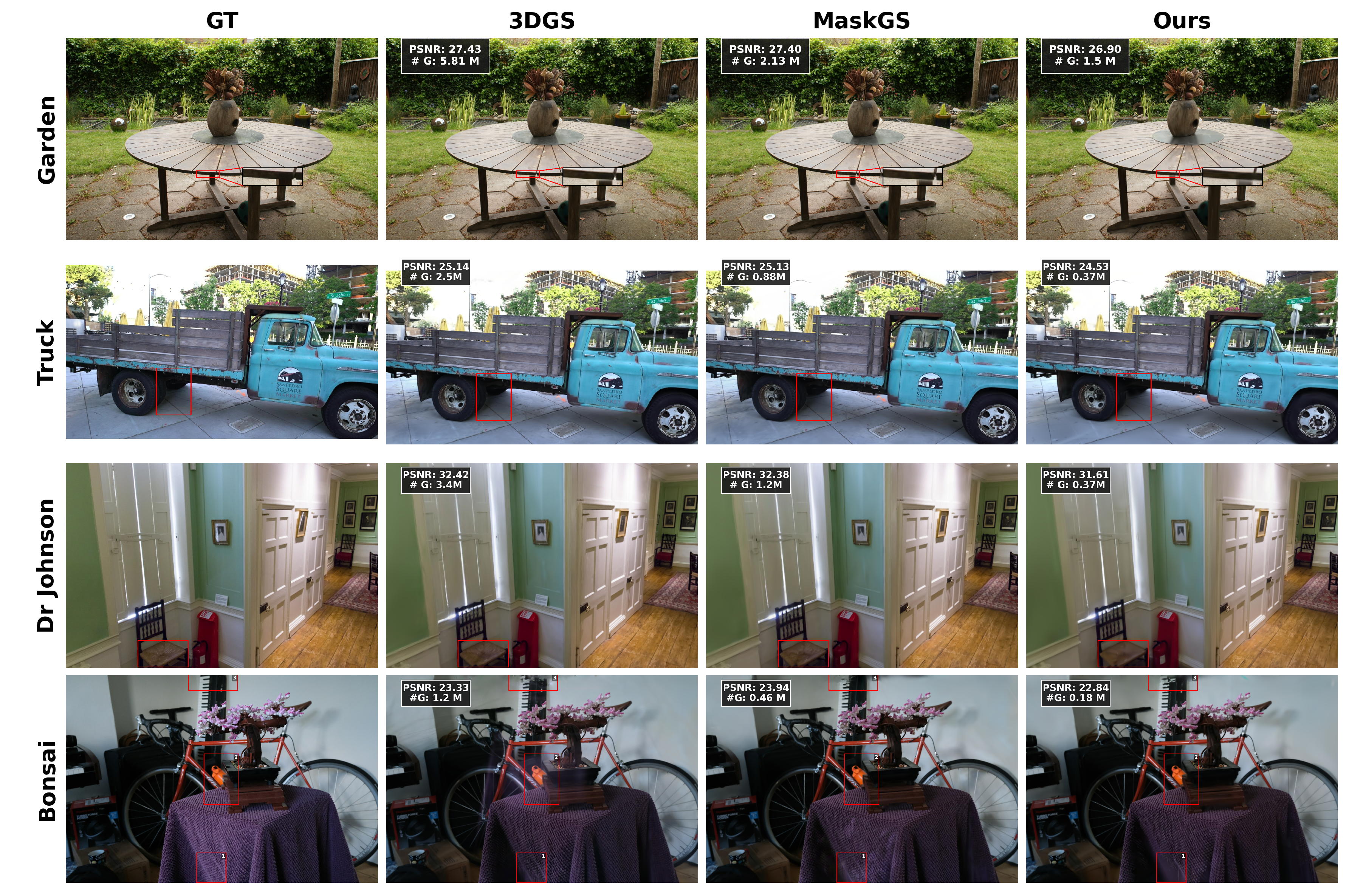}
\caption{Qualitative comparison of our method with 3DGS and MaskGS. For each scene, we report PSNR (dB, $\uparrow$) and \#G (M, $\downarrow$), where \#G denotes the number of Gaussians in millions. Regions with differences are marked in red boxes.}

  \label{fig:qualitative_results}
\end{figure*}

\noindent Equation~\eqref{eq:dF_dMi} decomposes the gradient into two interpretable components along the ray. The self term, \(1-\alpha_i T_i\), the amount of contribution of each Gaussian: when a Gaussian is faint or occluded, \(\alpha_i T_i\) is small, so \(1-\alpha_i T_i\) is large. In that case, increasing its mask would raise \(F\) the most, meaning the spatial mask assigns a stronger penalty to such \textit{low-importance} Gaussians. The occlusion term, \(\tfrac{\alpha_i}{\,1-\alpha_i M_i\,}\sum_{j>i} M_j \alpha_j T_j\), captures the influence of \(i\) on all the Gausians behind it. If \(i\) has non-trivial opacity and there is a long, active tail of Gaussians deeper on the ray, this sum is large: keeping \(i\) occludes many others, so the gradient grows and discourages \(M_i\). If small Gaussians lie behind, this occlusion term is small. Together, these effects form a simple behavior: weak or occluded Gaussians receive a stronger push toward smaller masks, while strong foreground Gaussians—especially when little lies behind—see a much milder signal and are preserved.

Beyond these per-Gaussian effects, the same decomposition predicts how the gradient scales with the overall density of active masks. The gradient becomes stronger when a larger share of masks are active. Let \(p\) denote the probability that a mask is “on”. As \(p\) increases, more Gaussians participate along each ray and transmittance \(T_i\) becomes smaller for deeper indices, which enlarges the self term \(1-\alpha_i T_i\), and the interaction term grows because the tail sum \(\sum_{j>i} M_j \alpha_j T_j\) is larger. This yields larger \(|\partial F/\partial M_i|\) in high-density settings than in sparse ones. Thus, the signal adapts to the \emph{population} of active masks: it pushes harder when many masks are on (e.g., \(p\!\approx\!0.8\)) and more gently when few are on (e.g., \(p\!\approx\!0.1\)).

\subsection{Spatial Mask Loss}
Given the \emph{spatial mask} from Eq.~\eqref{eq:redundancy-map}, let \(x=(u,v)\) denote a pixel on the \(H\times W\) image grid and write \(F(x)\equiv F(u,v)\).
We define the per-view mask loss as the mean energy of this spatial mask:
\begin{equation}
\label{eq:mask-loss-mean-energy}
L_{\text{mask}}
\;=\;
\frac{1}{HW}\sum_{u=1}^{H}\sum_{v=1}^{W} F(u,v)^{2}
\;
\end{equation}

\noindent The total objective combines RGB reconstruction loss with this regularizer:
\begin{equation}
\label{eq:total-loss}
L \;=\; L_{\text{rgb}} \;+\; \lambda_{F}\, L_{\text{mask}}\,,
\end{equation}
where \(\lambda_{F}\!\ge 0\) is a scalar hyperparameter that controls the strength of the spatial-mask penalty. For pruning, we follow the exact setup as MaskGS\cite{maskgs} explained in Section~\ref{subsec:preliminaries_maskgs}


\section{Experiments}

\subsection{Dataset and Metrics}
\label{sec:dataset}

We conduct experiments on three benchmark datasets. 
The Tanks\&Temples dataset~\cite{TT_dataset} provides outdoor scenes, 
while the Deep Blending dataset~\cite{DB_dataset} focuses on indoor environments. 
In addition, we include Mip\hbox{-}NeRF360~\cite{mip_dataset}, which contains both indoor and outdoor scenes.
We evaluate performance using Peak Signal-to-Noise Ratio (PSNR)~\cite{psnr}, which measures pixel-level differences; 
structural similarity (SSIM)~\cite{SSIM}, which reflects consistency of luminance, contrast, and structure; 
and learned perceptual image patch similarity (LPIPS)~\cite{lpips}, which compares image features in a neural network space. 
We also report the number of Gaussians (\#GS).

\subsection{Compared Baselines}
\label{sec:baseline}
We compare our approach with four methods. The original 3DGS~\cite{3dgs_orig} serves as the primary reference. LightGaussian~\cite{lightgaussian} uses score-based pruning followed by a standard recovery phase (continued optimization without densification or additional pruning), a practice we also adopt. Compact3DGS~\cite{compact3dgs} prunes via a learned volume mask, and MaskGS~\cite{maskgs} employs probabilistic masks with masked rasterization.

\subsection{Implementation Details}
We integrate our method into the MaskGS~\cite{maskgs} rasterizer. Since the spatial-mask computation and its gradients are tightly coupled to the renderer’s per-splat compositing, there is no practical way to implement them in Python; we therefore implement both parts inside the CUDA rasterizer. In the forward pass, the spatial mask \(F\) is computed alongside standard RGB compositing; in the backward pass, we compute the gradients in CUDA and make them available to PyTorch for backpropagation.
For all methods (ours and baselines), each scene is trained for 30K steps, followed by a 5K step recovery phase that optimizes only the 3DGS parameters (no densification or pruning), as in LightGaussian~\cite{lightgaussian}. All results in Table~\ref{tab:cross_dataset_comparison} use this 30k\,+\,5k schedule.

\subsection{Quantitative Results}

\noindent\textbf{Tanks\&Temples (outdoor).}
Compared to MaskGS, our method reduces \#GS by \textbf{2.2}\,$\times$ (from \(0.590\,\mathrm{M}\) to \(0.272\,\mathrm{M}\)); relative to 3DGS, the reduction is \textbf{6.7}\,$\times$ (from \(1.825\,\mathrm{M}\) to \(0.272\,\mathrm{M}\)). Quality changes are small: PSNR \(23.25\) vs.\ \(23.75\) (\(-0.50\) dB), SSIM \(0.829\) vs.\ \(0.847\) (\(-0.018\)), LPIPS \(0.221\) vs.\ \(0.178\) (\(+0.043\)). In other words, the reduction in \#GS is substantial, but the loss in image quality is comparatively small. Relative to the original 3DGS baseline, we achieve a \textbf{6.7}\,$\times$ reduction in \#GS (from \(1.825\,\mathrm{M}\) to \(0.272\,\mathrm{M}\)). The qualitative results (Section~\ref{subsec:qualitative}) provide visual context for these differences.

\noindent\textbf{Deep Blending (indoor).}
We obtain a reduction in \#GS of \(\approx 2.7\times\) fewer (from \(0.694\,\mathrm{M}\) to \(0.255\,\mathrm{M}\)) relative to MaskGS, with small changes in quality: PSNR \(29.46\) vs.\ \(29.71\) (\(-0.25\) dB), SSIM \(0.904\) vs.\ \(0.905\) (\(-0.001\)), and LPIPS \(0.261\) vs.\ \(0.246\) (\(+0.015\)). The near-identical SSIM suggests structural details are preserved despite aggressive pruning. Compared to the original 3DGS baseline, the reduction is \(\approx 11.0\times\) fewer (from \(2.815\,\mathrm{M}\) to \(0.255\,\mathrm{M}\)). This reduction lowers both on-disk model size and GPU memory during rendering, and it reduces per-frame compositing cost. We use the same hyperparameters across scenes, and observe similar reductions without dataset-specific tuning.
\begin{figure*} 
  \centering
  \includegraphics[width=\textwidth]{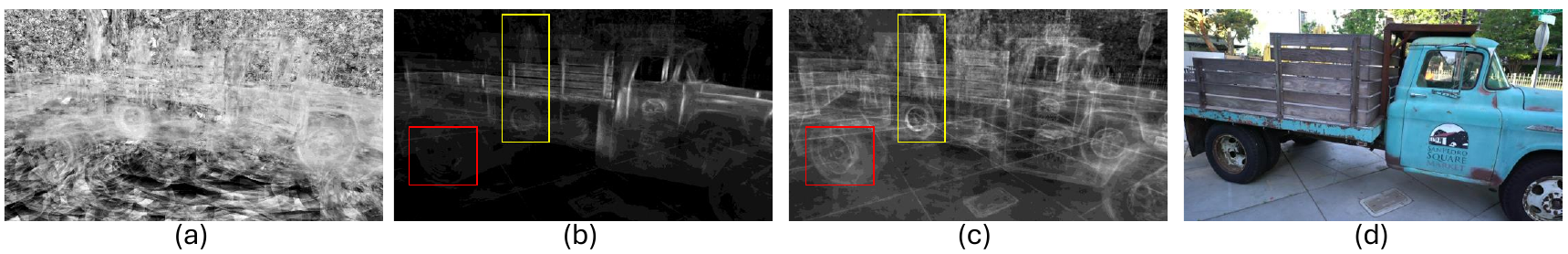}
\caption{\textbf{Effect of the forward function on the spatial mask.}
We visualize the spatial mask $F(x)$ produced by three forward designs; brighter intensities indicate higher penalty / lower importance.
(a) \emph{Scenario A} (Eq.~\eqref{eq:scenarioA_weights_FA}): the inverse-importance weights $w_i = 1/(\alpha_i T_i+\varepsilon)$ explode where $\alpha_i T_i\!\ll\!1$, causing saturation and broad regions to be marked as \textit{low-importance}.
(b) \emph{Scenario B} (Eq.~\eqref{eq:scenarioB_forward}): the cumulative-transmittance term $1-T_i$ emphasizes occluded content (yellow box) but tends to over-penalize with depth and under-respond to faint foreground structure (red box).
(c) \emph{Proposed} spatial mask (Eq.~\eqref{eq:redundancy-map}): the factor $1-\alpha_i T_i$ highlights occluded areas (yellow) while also capturing non-occluded, low-opacity regions (red) with reduced background noise.
(d) Reference RGB image.}
  \label{fig:various_forwards}
\end{figure*}

\noindent\textbf{Mip\hbox{-}NeRF360 (indoor/outdoor).}
On this mixed benchmark, \#GS decreases by \(\approx 1.4\times\) fewer (from \(1.205\,\mathrm{M}\) to \(0.866\,\mathrm{M}\)) versus MaskGS. The quality gap is moderate: PSNR \(26.75\) vs.\ \(27.49\) (\(-0.74\) dB), SSIM \(0.800\) vs.\ \(0.811\) (\(-0.011\)), and LPIPS \(0.255\) vs.\ \(0.228\) (\(+0.027\)). Relative to the original 3DGS baseline, the reduction is \(\approx 3.7\times\) fewer (from \(3.204\,\mathrm{M}\) to \(0.866\,\mathrm{M}\)). These outcomes are consistent with the challenge of \emph{unbounded} outdoor views in Mip\hbox{-}NeRF360—scenes that cannot be enclosed within a small finite box around the cameras (e.g., with sky or distant geometry). We use the same hyperparameters across all datasets for fairness.

\subsection{Qualitative Results}
\label{subsec:qualitative}
\noindent\textbf{Qualitative comparisons.} We present visual comparisons for novel-view synthesis in Fig.~\ref{fig:qualitative_results}. In the top row (Garden), the PSNR difference between our method and 3DGS is \(0.53\,\mathrm{dB}\); the only visible change is a slightly blurred edge on the table, which is hard to notice without zooming. In terms of \#GS, our method uses \(1.42\times\) fewer Gaussians than MaskGS and \(3.87\times\) fewer than 3DGS. Overall, the quality difference is small compared with the substantial reduction in the number of Gaussians.

The second row of Fig.~\ref{fig:qualitative_results}, shows a scene from the Tanks\&Temples dataset. The PSNR gap between our method and 3DGS (which achieves the highest PSNR for this scene) is \(0.61\,\mathrm{dB}\). The main visual difference is that the partially occluded rear wheel of the truck appears slightly less sharp in our result compared to 3DGS and MaskGS. However, this sharper detail comes at a cost: our method uses \(2.38\times\) fewer Gaussians than MaskGS and \(6.76\times\) fewer than 3DGS. Thus, the loss in visual quality is minor relative to the substantial reduction in the number of Gaussians.

Third row of Fig.~\ref{fig:qualitative_results}, shows a scene from the Deep Blending dataset (Dr Johnson). The PSNR gap between our method and MaskGS is \(0.77\,\mathrm{dB}\). As highlighted with red box in the rendered images, the only noticeable difference is a slight variation in the chair’s texture color between the two methods. This accounts for the \(0.77\,\mathrm{dB}\) difference, while no meaningful visual degradation---such as deformation, missing regions, or other artifacts---can be observed in the rendered images. Importantly, our method achieves this result with only \(0.37\)M Gaussians, which is \(3.24\times\) fewer than MaskGS (1.2M) and \(9.19\times\) fewer than 3DGS (3.4M).

In the fourth row of Fig.~\ref{fig:qualitative_results}, we present an extreme case where the PSNR difference between our method and MaskGS is \(1.10\,\mathrm{dB}\). As highlighted by the red boxes: in box 2, both our result and MaskGS render the region cleanly, whereas 3DGS shows slight blur; in box 3, none of the methods recovers the keyboard edge precisely; and in box 1, the only notable difference between our result and MaskGS is a subtle fade in the fabric texture. Despite these local deviations, our rendering uses \(2.56\times\) fewer Gaussians than MaskGS and \(6.67\times\) fewer than 3DGS, representing a substantial reduction in \#GS for a comparatively small change in image quality.

\begin{table}[t]
  \centering
  \setlength{\tabcolsep}{10pt}
  \renewcommand{\arraystretch}{1.1}
    \caption{Ablation of forward designs for the spatial mask on Tanks\&Temples ( \#GS denotes millions of Gaussians)}
\label{tab:forward_mask_ablation_tandt}
  \begin{tabular}{l | c c c c}
    \toprule
    Method & PSNR$\uparrow$ & SSIM$\uparrow$ & LPIPS$\downarrow$ & \#GS$\downarrow$ \\
    \midrule
    Scenario A & 21.12 & 0.774 & 0.253 & 0.365 \\
    Scenario B  & 22.82 & 0.803 & 0.239 & 0.471 \\
    Ours    & \cellcolor{best}23.25 & \cellcolor{best}0.829 & \cellcolor{best}0.221 & \cellcolor{best}0.272 \\
    \bottomrule
  \end{tabular}
\end{table}

\subsection{Effect of $\lambda_F$}
Increasing $\lambda_F$ scales the spatial-mask penalty in Eq.~\eqref{eq:total-loss}, emphasizing
$L_{\text{mask}}=\tfrac{1}{HW}\sum_{u,v}F(u,v)^2$. This drives the gradients of \emph{low-importance} Gaussians (small $\alpha_i T_i$) toward zero, inducing earlier and stronger pruning (as in Section.~\ref{subsec:preliminaries_maskgs}).

\begin{figure}
  \centering
\includegraphics[width=0.95\columnwidth]{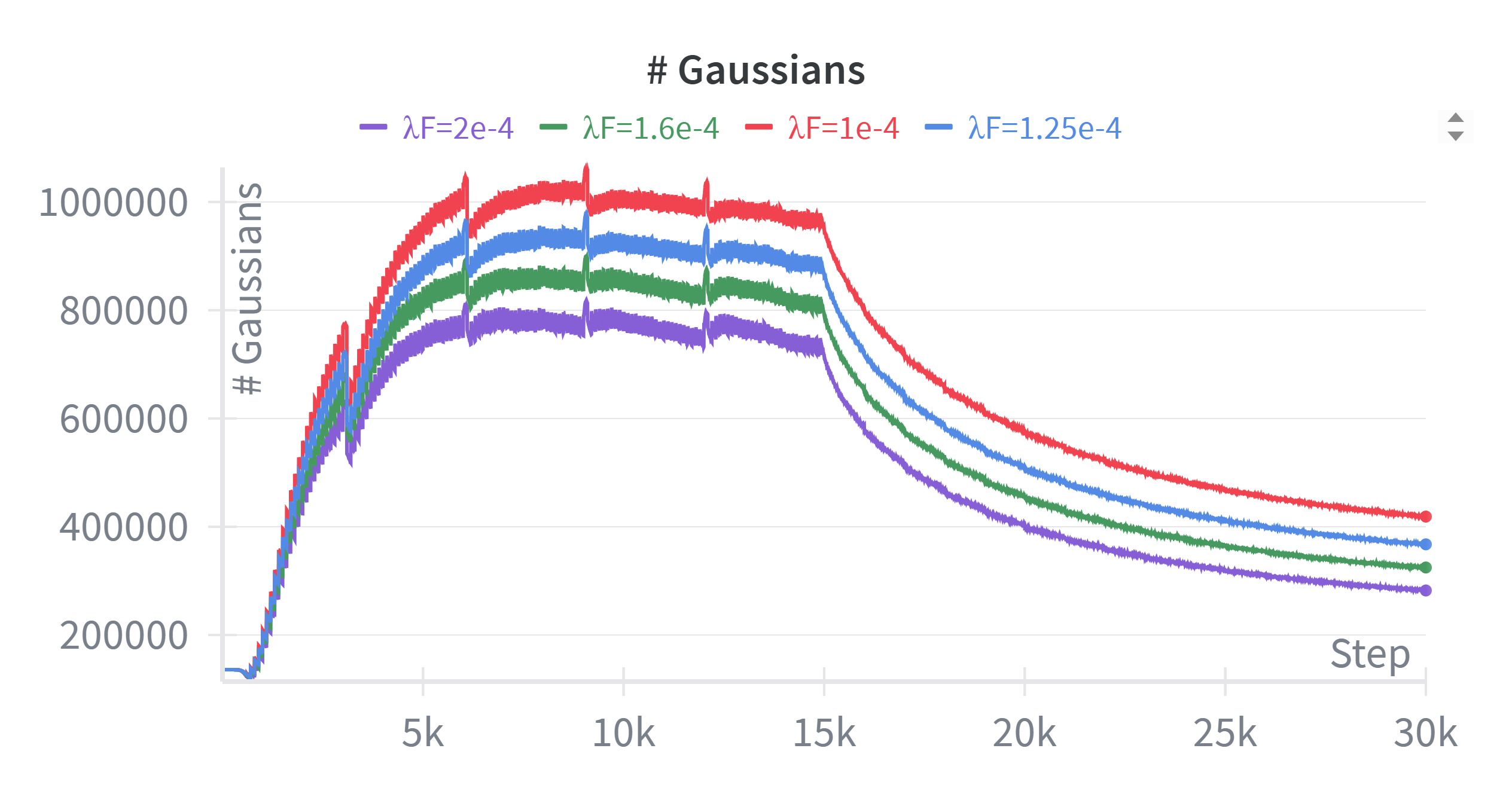} 
\includegraphics[width=0.95\columnwidth]{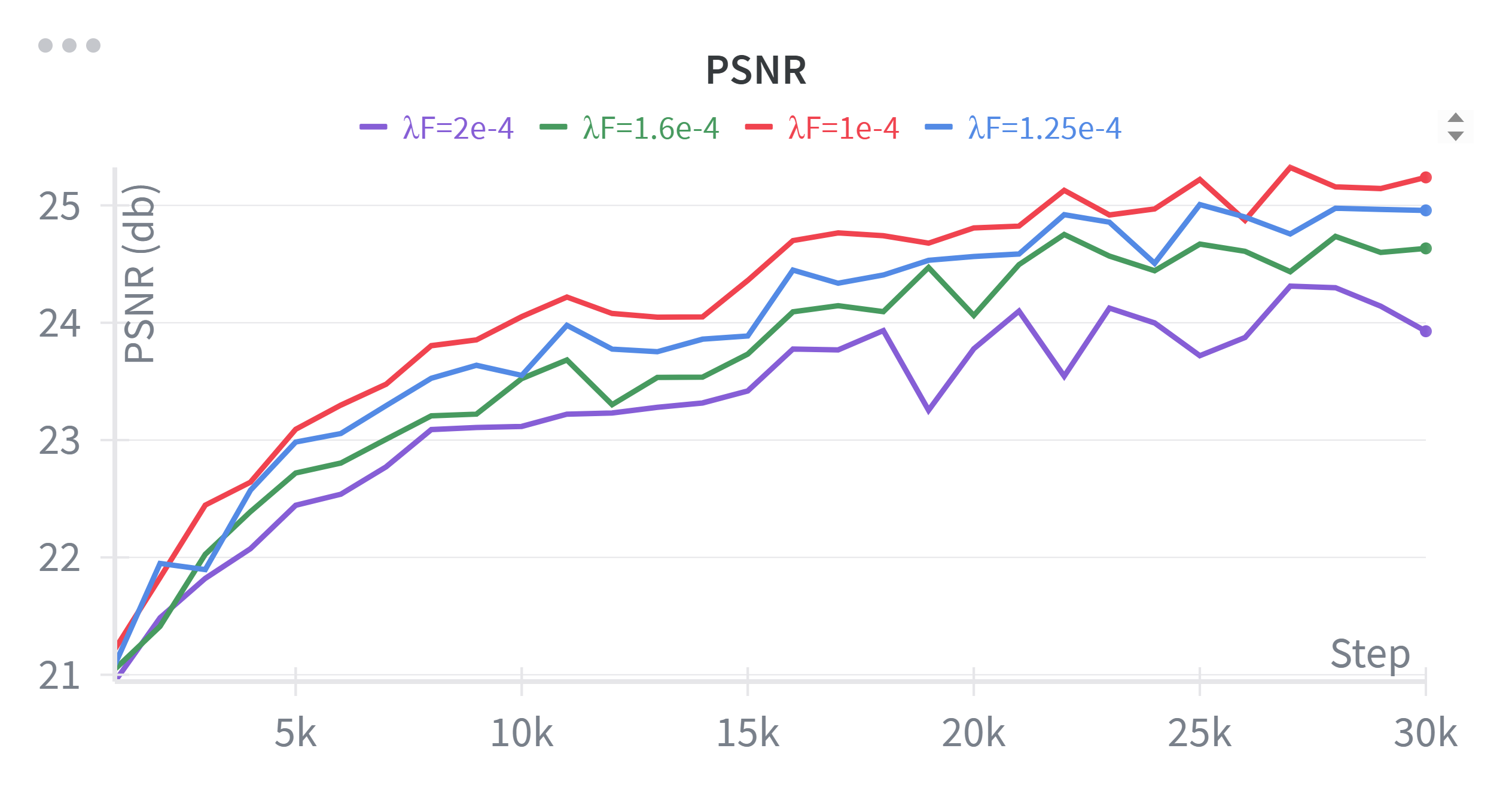} 
\caption{Effect of $\lambda_F$ on number of Gaussians (top) and quality (bottom). Larger $\lambda_F$ prunes more and ends with fewer Gaussians, but PSNR drops.}
\label{fig:wandb}
\end{figure}

We sweep $\lambda_F \in \{1,\,1.25,\,1.6,\,2\}\!\times\!10^{-4}$. As shown in Fig.~\ref{fig:wandb}, a larger $\lambda_F$ consistently yields fewer Gaussians throughout training and a lower final count, with a corresponding PSNR drop. $\lambda_F\!=\!1\!\times\!10^{-4}$ gives the best PSNR; $\lambda_F\!=\!1.25\!\times\!10^{-4}$ is close (about $0.3$--$0.5$\,dB lower) while using $\sim\!8$--$12\%$ fewer Gaussians. Higher values ($1.6$--$2\!\times\!10^{-4}$) prune more but incur a $\sim\!0.7$--$1.2$\,dB PSNR loss.

\subsection{Ablations on the Forward Design}
Designing the forward mask aggregation is critical because it determines the signal our loss will propagate. We did not assume the final form a priori; instead, we treated it as a hypothesis and tested two other alternatives that emphasize low-importance Gaussians. Each variant was implemented in CUDA for a fair comparison; see Fig.~\ref{fig:various_forwards}, and quantitative results are reported in Table~\ref{tab:forward_mask_ablation_tandt}. We adopt Eq.~\eqref{eq:redundancy-map} because it is stable, aligns with alpha compositing, and provides clean, importance-aware gradients. Below, we summarize the two alternatives and explain why they fall short.

\subsubsection{Scenario A: Inverse-importance weighting}
Here, we treat the \emph{inverse} of per-Gaussian importance, \(1/(\alpha_i T_i)\), as the low-importance signal, so Gaussians with small \(\alpha_i T_i\) receive larger weight. We keep the transmittance:
\begin{equation}
\label{eq:scenarioA_forward_exact}
T_0 = 1,
\qquad
T_{i+1} = (1-\alpha_i M_i)\,T_i,
\end{equation}
and define the weights and normalized aggregation
\begin{equation}
\label{eq:scenarioA_weights_FA}
w_i \;=\; \frac{1}{\alpha_i T_i + \varepsilon},
\qquad
F_A \;=\; \frac{\sum_i w_i M_i}{\sum_i w_i}.
\end{equation}
For shorthand, let
\begin{equation}
\label{eq:scenarioA_defs}
S \coloneqq \sum_k w_k,
\qquad
W^{>}_i \coloneqq \sum_{j>i} w_j,
\qquad
F^{>}_i \coloneqq \sum_{j>i} w_j M_j.
\end{equation}
The gradient with respect to \(M_i\) can be written as
\begin{equation}
\label{eq:scenarioA_backward_exact}
\begin{aligned}
\frac{\partial F_A}{\partial M_i}
&= \frac{1}{S}\Bigl[w_i + \sum_{j>i} (M_j - F_A)\,\frac{\partial w_j}{\partial M_i}\Bigr],\\[2pt]
\frac{\partial w_j}{\partial M_i}
&= \frac{\alpha_i \alpha_j T_j}{(1 - \alpha_i M_i)\,(\alpha_j T_j + \varepsilon)^2}.
\end{aligned}
\end{equation}

\noindent Scenario A falls short because the inverse-importance weights \(w_i=1/(\alpha_i T_i+\varepsilon)\) explode when \(\alpha_i T_i\) is tiny, saturating \(F_A\) and tagging broad regions as low-importance (Fig.\ref{fig:various_forwards} (a)). As a result, the mask overreacts to near-zero contributions and background noise instead of reliably isolating truly low-important Gaussians.

\subsubsection{Scenario B: Cumulative-transmittance masking}
Here, we suppress each Gaussian by the cumulative occlusion in front of it. With the standard
transmittance,
\begin{equation}
\label{eq:scenarioB_forward}
\begin{aligned}
T_0 &= 1, &\qquad T_{i+1} &= (1-\alpha_i M_i)\,T_i,\\
f_i &\coloneqq M_i(1 - T_i), &\qquad
F_B &= \frac{1}{\log(1+N)}\sum_i f_i .
\end{aligned}
\end{equation}

\noindent The gradient w.r.t.\ \(M_i\) is
\begin{equation}
\label{eq:scenarioB_backward_exact}
\frac{\partial F_B}{\partial M_i}
= \frac{1}{\log(1+N)}
\left[
(1 - T_i)
+ \frac{\alpha_i}{1 - \alpha_i M_i}\,
\sum_{j>i} M_j\,T_j
\right].
\end{equation}

\noindent Scenario B falls short because the cumulative term $(1 - T_i)$ depends only on occlusion, so the mask primarily tracks how far along the ray a Gaussian sits rather than its per-Gaussian importance. This depth bias over-penalizes long rays and occluded regions while under-responding to faint foreground structure (Fig.~\ref{fig:various_forwards} (b)).

We implemented all three scenarios in CUDA. The spatial masks are shown in Fig.\ref{fig:various_forwards}. As seen in the 
Table~\ref{tab:forward_mask_ablation_tandt}, Scenario A produces fewer Gaussians than Scenario B (\#GS 0.365M vs.\ 0.471M) but suffers a 1.7\,dB PSNR drop (21.12 vs.\ 22.82) because its inverse-importance weighting over-penalizes background Gaussians, saturating the mask (Fig.~\ref{fig:various_forwards}(a)). By contrast, the proposed forward achieves the highest pruning rate (fewest Gaussians; \#GS 0.272M) while also delivering the best reconstruction quality (highest PSNR/SSIM and lowest LPIPS).

\section{CONCLUSIONS}
This work introduces \textbf{SVR-GS}, a spatially variant regularizer for probabilistic mask pruning in 3D Gaussian Splatting. It forms a local per-pixel mask from each Gaussian’s visibility-weighted contribution, aligning sparsity with the reconstruction loss. A CUDA ablation design motivated our mask aggregation. Integrated into masked rasterization, and averaged over three real-world datasets, SVR-GS prunes Gaussians by \textbf{1.79}\,$\times$ vs.\ MaskGS and \textbf{5.63}\,$\times$ vs.\ 3DGS, with PSNR drops of at most \textbf{0.5}\,dB (\textbf{0.50} and \textbf{0.40}\,dB). These reductions shrink disk and GPU memory footprints and accelerate inference, supporting real-time robotics use cases.









\bibliographystyle{IEEEtran}
\bibliography{root}  

\end{document}